\documentclass{article}

\usepackage[preprint]{neurips_2024}

\usepackage[utf8]{inputenc} % allow utf-8 input
\usepackage[T1]{fontenc}    % use 8-bit T1 fonts
\usepackage{hyperref}       % hyperlinks
\usepackage{url}            % simple URL typesetting
\usepackage{booktabs}       % professional-quality tables
\usepackage{amsfonts}       % blackboard math symbols
\usepackage{nicefrac}       % compact symbols for 1/2, etc.
\usepackage{microtype}      % microtypography
\usepackage{xcolor}         % colors
\usepackage{mathptmx}
\usepackage{amsmath}
\usepackage{graphicx}
\usepackage{algorithm}
\usepackage{algorithmic}

\usepackage{multirow}
\usepackage{bm}
\newcommand{\jr}{\textcolor{black}}

\title{LLplace: The 3D Indoor Scene Layout Generation and Editing via Large Language Model}

\author{
  Yixuan Yang\textsuperscript{1,2} \thanks{Email: arnoldyang97@gmail.com.} \qquad  Junru Lu\textsuperscript{2}\qquad Zixiang Zhao\textsuperscript{3}\qquad Zhen Luo\textsuperscript{1} \AND James J.Q. Yu\textsuperscript{4}\qquad Victor Sanchez\textsuperscript{2}\qquad Feng Zheng\textsuperscript{1}\thanks{Corresponding Author.} \\
\\
  \textsuperscript{1} Southern University of Science and Technology\qquad \textsuperscript{2} University of Warwick\\  \textsuperscript{3}Xi'an Jiao Tong University\qquad \textsuperscript{4}York University \\
}

\begin{document}

\maketitle

\begin{abstract}
Designing 3D indoor layouts is a crucial task with significant applications in virtual reality, interior design, and automated space planning. Existing methods for 3D layout design either rely on diffusion models, which utilize spatial relationship priors, or heavily leverage the inferential capabilities of proprietary Large Language Models (LLMs)
, which require extensive prompt engineering and in-context exemplars via black-box trials. These methods often face limitations in generalization and dynamic scene editing. In this paper, we introduce LLplace, a novel 3D indoor scene layout designer based on lightweight fine-tuned open-source LLM Llama3. LLplace circumvents the need for spatial relationship priors and in-context exemplars, enabling efficient and credible room layout generation based solely on user inputs specifying the room type and desired objects. We curated a new dialogue dataset based on the 3D-Front dataset, expanding the original data volume and incorporating dialogue data for adding and removing objects. This dataset can enhance the LLM’s spatial understanding. 
Furthermore, through dialogue, LLplace activates the LLM's capability to understand 3D layouts and perform dynamic scene editing, enabling the addition and removal of objects.
Our approach demonstrates that LLplace can effectively generate and edit 3D indoor layouts interactively and outperform existing methods in delivering high-quality 3D design solutions.
Code and dataset will be released.
\end{abstract}

\section{Introduction}
The design and optimization of \jr{3D} indoor object layouts play a crucial role in various applications, including interior design~\cite{feng2024layoutgpt,paschalidou2021atiss}, game design~\cite{deitke2022️ProcTHOR}, automated space planning~\cite{fu2024scenellm,huang2023leo}, and robotics~\cite{yang2024physcene,chen2023ll3da}. 
Effective and reasonable indoor layout design enhances both the functionality and aesthetic appeal of living and working spaces, directly impacting the quality of life and productivity of their occupants. 
Despite significant advancements in the field of artificial intelligence, specifically in natural language processing and computer vision, the task of \jr{flexibly} generating and dynamically editing 3D indoor layouts \jr{from naive texts} remains a complex challenge.

Existing methods for designing indoor scene layouts are primarily categorized into two %ways
\jr{types}. 
The first one is based on diffusion models\,\cite{ho2020denoising}, which utilize these models along with various spatial feature priors to generate 3D layouts. 
Representative %models
\jr{approaches} in this category include DiffuScene~\cite{tang2023diffuscene} and InstructScene~\cite{lin2024instructscene}. 
The second category relies on the inferential capabilities of existing %GPT models
\jr{LLMs (e.g., GPT-4~\cite{openai2024gpt4})}, using numerous prompts to generate corresponding 3D layouts, such as LayoutGPT~\cite{feng2024layoutgpt} and Holodeck~\cite{yang2024holodeck}.
DiffuScene~\cite{tang2023diffuscene} extracts the multidimensional features of objects in space and uses diffusion to achieve the %training
\jr{self-denosing learning} and generation of 3D spatial layouts. In contrast, InstructScene~\cite{lin2024instructscene} leverages the positional relationships between objects as conditions, constructing a graph model where each node represents an object,
\jr{then }uses graph diffusion to generate the layout.
On the other hand, %GPT-based
\jr{LLM-based} methods differ from diffusion-based models\jr{, as they utilize inherent language understanding and generation capabilities to interpret textual prompts and translate them into spatial arrangements}. Holodeck requires %users to provide 
specific spatial relationships between objects as prompts to generate the room layout, while LayoutGPT %uses user inputs to 
first retrieve\jr{s} relevant room layouts from a \jr{well-crafted} database and then use these as in-context exemplars to guide the %GPT model
\jr{LLMs }in generating the \jr{targeting} layout.
In summary, %each existing method has its drawbacks.
\jr{above existing approaches have obvious drawbacks.} 
Firstly, most layout generation models rely on spatial relationship priors or examples as model inputs to guide in generation. 
If users do not provide these relationships or if the system cannot retrieve accurate examples, these models cannot achieve convincing results.
%This also significantly impacts the model’s generalization capability and its capacity to adapt to different scenarios. 
\jr{Herein, these prior-inspired strategies significantly constrain the model's generalization capability when meets newly different scenarios, where high-quality priors or exemplars are expensive.}
Secondly, most current \jr{LLM-based layout} models %can statically complete layout generation under guidance but are overly dependent on graph relationships and examples, and cannot perform dynamic scene editing. 
\jr{solely supports one-time static layout generation, while can not perform dynamic scene editing.}
This does not align with the interactive nature intended for %large language models (
LLMs%)
.
Therefore, we %aim to propose a dynamic scene layout designer that does not depend on strong priors or in-context exemplars and can interactively use large language models.
\jr{are particularly interested in exploring the potential of LLMs as a dynamic 3D scene layout designer that do not rely on strong priors or pre-prepared in-context exemplars.} 

\begin{figure}[!t]
\vspace{-1em}
	\centering
	\includegraphics[width=0.98\linewidth]{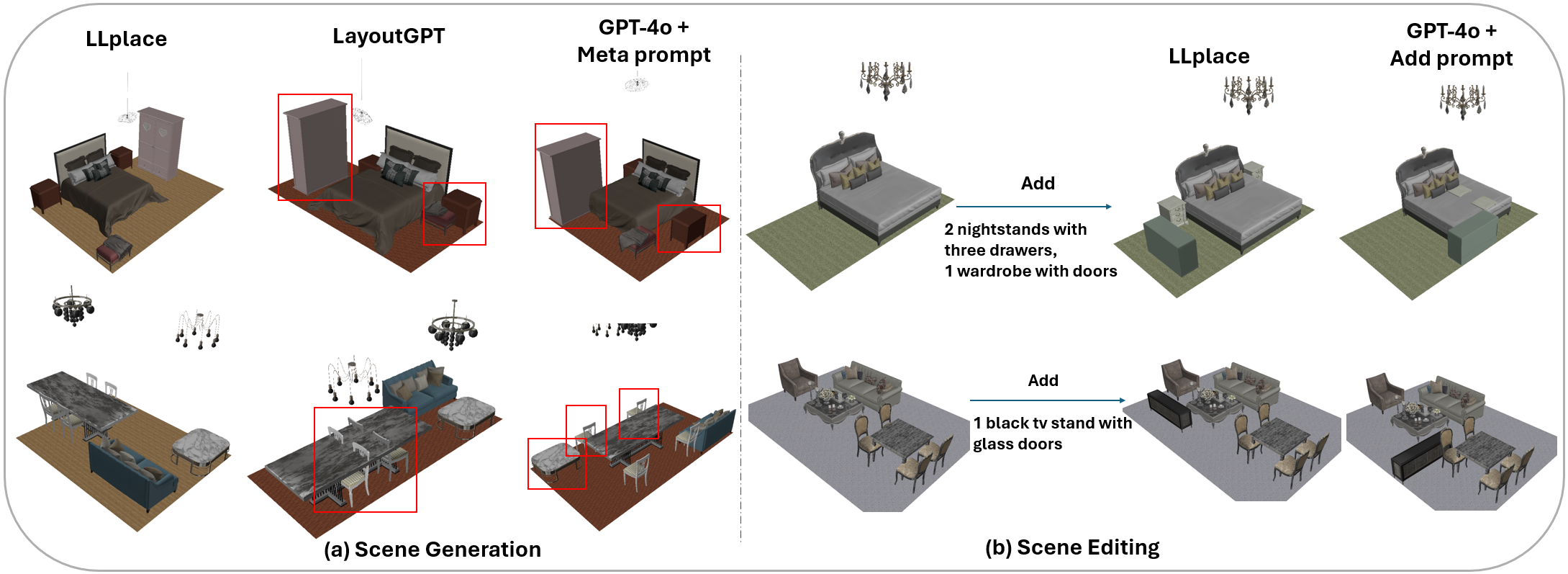}
 \vspace{-1em}
	\caption{Generated results of 3D indoor scenes from LLplace and compared with \emph{LayoutGPT} and \emph{GPT-4o}. And editing results from LLplace and compared with \emph{GPT-4o}. 
}
\label{first image}
\vspace{-2em}
\end{figure}

%To address the above issues,
\jr{Consequently}, in this paper, we introduce a novel 3D indoor scene layout designer, LLplace (\textbf{L}arge \textbf{L}anguage Model for Indoor \textbf{Place}ment). %Our designer is built on a fine-tuned Llama3 model and includes a comprehensive set of instruction prompts. 
\jr{We first carefully design a format-friendly meta prompt template towards 3D indoor layout design, then reconstruct regular 3D-Front dataset\,\cite{3d-front} for static scene generation and dynamic scenario editing in the format of multi-turn conversations. This is to ensure compatibility with the interactive routines of LLMs.}
%Additionally, we extend the 3D-Front dataset with the dialogue data to train the designer's conversational capabilities and activate dynamic scene editing capability for LLM.
Specifically,
%due to the closed-nature nature of GPT models, indoor design models based on GPT require highly complex prompt designs, often necessitating intricate templates and multiple positional relationships between objects or in-context examples to guide the completion. Moreover, due to the static nature of the generation process, it is often hard to continue editing the scene layout afterward.
we opt to fine-tune a \jr{SOTA} open-source LLM, Llama3~\cite{touvron2023llama}.%, to \jr{simultaneously} accomplish %specific \jr{static} layout generation and scene editing tasks.
%To facilitate training, 
We employ LoRA~\cite{hu2021lora} for parameter-efficient fine-tuning.
In our design pipeline, we first specify the user input as the room type and descriptions of objects within the room. We then retrieve 3D assets and corresponding bounding boxes from the 3D-Front dataset using the object descriptions. 
Subsequently, we convert the user inputs and %the
\jr{corresponding} bounding boxes of the corresponding objects into a JSON format that the LLM can accept, which is like the general data process in the LLM area~\cite{liu2023alignbench,lu2023memochat}. The entire transcription is finally completed after we embed the user request JSON with our meta prompt template. The overall pipeline is not only used for the construction of training data, but also for the execution of inference.
\jr{In accordance with the input JSON format, we also design to use JSON to standardize the labels of the training data. The ``JSON-in'' and ``JSON-out'' schema is beneficial for the coupling of semi-structured natural language requests and auxiliary structured programming. Based on the retrieved 3D assets and their bounding boxes, we ask the LLM to report its design}
containing the coordinates and rotation angles of the objects in the room.
\jr{We go beyond traditional static 3D indoor layout generation by also considering dynamic scene editing. We develop the aforementioned instructions and labels into dialogues, adding an additional round of editing requests, such as adding or removing objects. The LLM then reasonably modifies its further output accordingly. In addition, we are able to refactor the user's input JSON and LLM's output JSON at each turn of the conversation}
into
spatial 3D bounding box layouts, which can then be rendered into a \jr{series of} 3D representations
.

As shown in Figure~\ref{first image}, we compare LLplace with LayoutGPT~\cite{feng2024layoutgpt} and the latest GPT-4o model with our meta prompt template. The scenes generated by LLplace are more reasonable compared to the other two models without overlapping and wrong rotation problems.
In scene editing, LLplace can understand the existing 3D scene and add objects to the correct positions. This demonstrates that the dynamic understanding and editing capabilities of the LLplace designer are not present in the existing LLM-based approaches.
The main contribution of the LLplace can be summarized as follows:
\begin{itemize}
\vspace{-0.7em}

\item We introduce a novel 3D indoor scene layout designer, LLplace, which is based on a fine-tuned open-source LLM model. This designer does not require the use of spatial relationship priors or in-context exemplars. Instead, it efficiently generates credible room layouts based solely on user inputs specifying the room type and the objects to be placed.

\item We curate a new dialogue dataset based on the 3D-Front dataset, which not only expands the original data volume but also includes dialogue data for adding and removing objects, enhancing the spatial understanding capabilities of the LLM \jr{towards the real physical world}.

\item By fine-tuned with this dialogue data, LLplace ensures that the LLM can statically generate 3D layouts. Also, it activates the LLM's capability to understand and generate 3D layouts via chatting, enabling the dynamic addition and removal of objects within the spatial layout.
\end{itemize}   

\section{Related Work}
Models for 3D indoor scene layout can be broadly classified into three categories: traditional methods using prior knowledge%for optimization
, generative models for scene generation, and LLM-based methods%for design
.

\noindent\textbf{Traditional 3D Indoor Scene Design.}
Early approaches to 3D indoor scene design used autoregressive models that required supervision with 2D bounding boxes or various visual maps ~\cite{ritchie2019fast,luo2020endtoend,yang2021indoor}.
And \cite{purkait2020vae,gao2023scenehgn,yang2021scene_uncentain} use variational auto-encoder (VAE)~\cite{VAE} to model the distribution of objects to generate indoor scenes.
SceneFormer~\cite{wang2021sceneformer} introduced the use of transformers to add furniture to scenes, marking a significant innovation in the field. Unlike previous methods that relied on separate models to predict different object attributes, ATISS~\cite{paschalidou2021atiss} demonstrated that a single transformer model could generate more realistic and efficient arrangements. 
%LEGO-NET~\cite{wei2023lego} further advanced the field by applying a transformer-based diffusion-like pipeline to refine coarse room layouts, learning human criteria for regularity. 
However, these traditional models often cannot use textual instructions to specify scene inputs and requirements.

\noindent\textbf{Generative Models for 3D Indoor Scene Design.}
Using diffusion models for indoor scene design has become increasingly popular~\cite{tang2023diffuscene,huang2023diffusionfor3d,fang2023ctrl,lin2024instructscene}. 
DiffuScene~\cite{tang2023diffuscene} extracts the features of various objects and uses a diffusion model to generate the characteristics of indoor scenes. 
Similarly, Ctrl-room~\cite{fang2023ctrl} generates bounding boxes and then employs ControlNet~\cite{zhang2023controlnet} to create panoramas, which are converted into textured meshes. InstructScene~\cite{lin2024instructscene} takes a different approach by utilizing relationships between objects as priors and applying graph diffusion for scene generation. Despite their strengths, diffusion-based models face challenges in achieving real-time interactivity and understanding existing scene layouts for further editing. 
These limitations make it difficult for diffusion models to facilitate dynamic scene modifications, which is a capability that shows potential in LLM-based models.

\noindent\textbf{LLM-Based Methods for 3D Indoor Scene Design.}
LLM-based methods leverage the inferential capabilities of LLMs to design 3D indoor layouts. These models use extensive prompts to generate corresponding layouts. 
For instance, Holodeck~\cite{yang2024holodeck} requires users to specify spatial relationships between objects as prompts to generate room layouts. 
Similarly, I-Design~\cite{ccelen2024design} first generates a graph of relationships, which is then used to create 2D design plans. 
~\cite{aguina2024open} represents an entire scene using a program, also requiring numerous prompts and spatial relationships to assist in generation. 
LayoutGPT~\cite{feng2024layoutgpt}, uses user inputs to retrieve relevant room layouts from a database and employs these as in-context exemplars to guide the GPT model in generating new layouts. 
While these methods demonstrate the capability of LLMs in layout design, they still face challenges in terms of flexibility due to their reliance on predefined examples. 
And also these methods do not utilize the inherent conversational capabilities of LLMs to further edit scenes after generation. This limits their interactivity and adaptability, as they cannot dynamically adjust layouts based on user dialogue.

\section{Method}
\begin{figure*}[!t]
\label{pipe}
	\centering
	\includegraphics[width=1\linewidth]{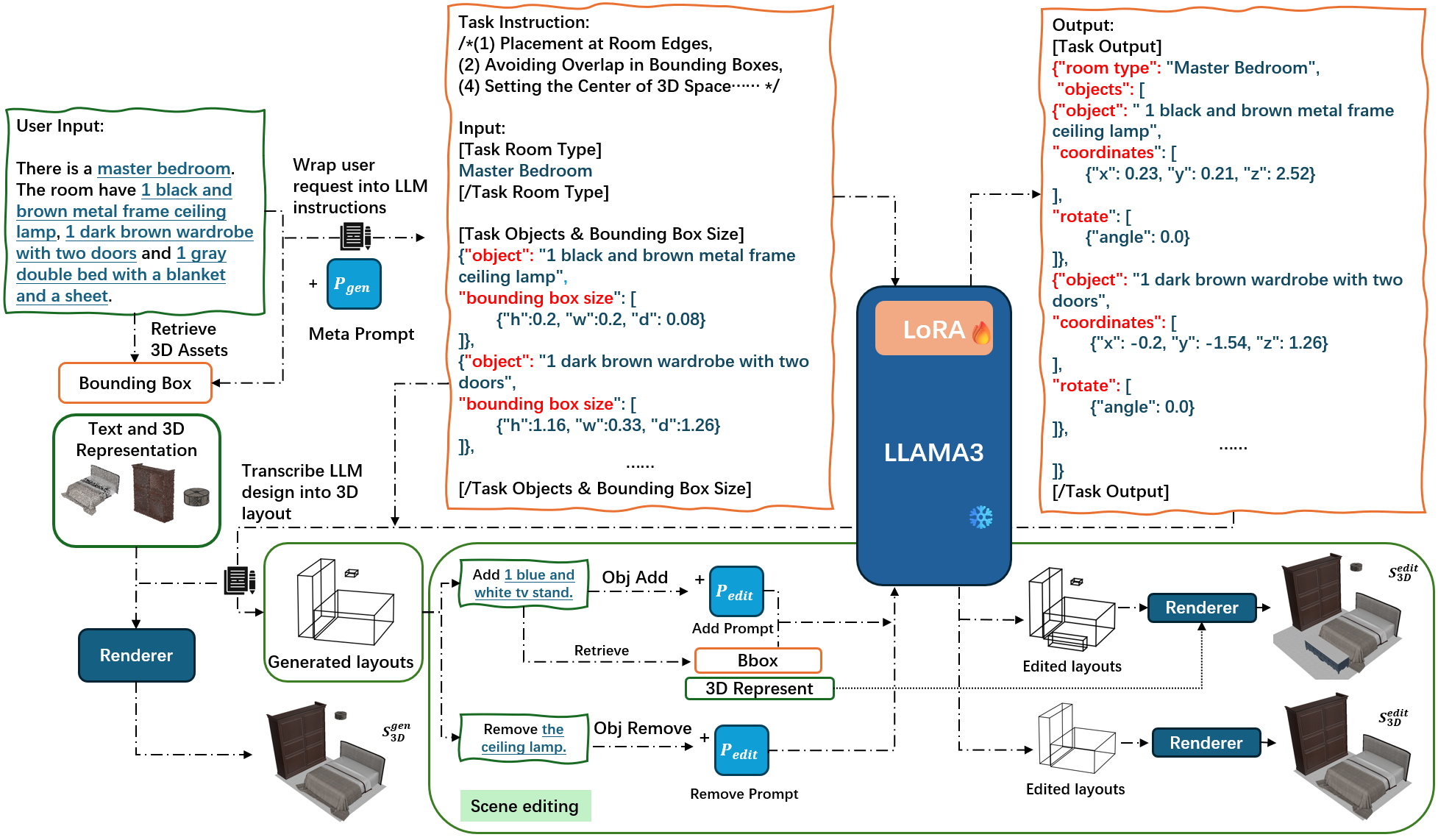}
 \vspace{-1em}
	\caption{The pipeline of LLplace. First, \jr{start from the left upper corner,} we extract the room type and the user's desired objects from the user input. Then, we retrieve 3D objects and corresponding bounding boxes. Next, we wrap the user input, bounding box information, and meta prompt $\bm{P}_{gen}$ into an LLM instruction, \jr{as shown in the middle upper box}. Using the LoRA fine-tuned Llama3 model, we obtain the LLM output \jr{(the right upper box)}, which includes the center coordinates and rotation angles of the objects. We then combine this output with the input information to convert it into a 3D layout and render it into a 3D scene \jr{(the left bottom corner)}. To edit the generated 3D scene layout, we combine the previous layout, user input, and edit prompt $\bm{P}_{edit}$ % to create a \jr{new editing} instruction. 
    \jr{into a new instruction.} The fine-tuned Llama3 model is then applied to generate the new scene, \jr{illustrated at the right bottom}.
}
\label{main_pipeline}
  \vspace{-1em}
\end{figure*}

The overall pipeline of LLplace is illustrated in Fig.~\ref{main_pipeline}. 
In section \ref{formulation}, we define the problem, highlighting our goal of not only generating room layouts but also enabling the LLM to understand layout distributions and perform layout editing.
Next, in section \ref{llplace}, we
\jr{present our approach, } proposing three strategies for model expansion. 
Then we detail how to define the input and output for the LLM model in section \ref{inputoutput} and explain how to construct effective \jr{meta} instruction prompts to guide layout generation and editing in section \ref{template}.
Finally, in section \ref{dialogue} we describe the process of constructing dialogue data that allows for layout generation and scene editing, based on the existing 3D-Front dataset.

\subsection{Problem Formulation} \label{formulation}
\vspace{-0.5em}
To support LLplace for designing indoor 3D scenes, the system imposes three requirements on user instruction $\mathcal{I}$: (1) The type of the room $\bm{T}$, (2) The %types of objects or 
specific descriptions $\bm{D}_{1 \sim N}$ of %the
\jr{$\bm{N}$ objects that} to be placed in the room, and (3) The specific quantity $\bm{Q}_n$ of the \jr{$\bm{n}$-th object}. 
$\bm{T}$ helps the designer to understand the overall objective of the room layout, clarifying whether the user intends to design a bedroom, living room, or another type of space.
Descriptions of the objects \jr{$\bm{D}_{1 \sim N}$} enable our designer to retrieve the most suitable items from an existing 3D database. 
Together with the specified quantities \jr{$\bm{Q}_n$} of each item, these descriptions allow the designer to design a practical room layout. Consequently, the user input can be structured as $\mathcal{I} = \left\{ \bm{T}, \left[ \bm{Q}_n, \bm{D}_n \right]_{1 \sim N} \right\}$%, where $\bm{n}$ is the number of objects in the room
.

Based on the user's description $\bm{D}_{1 \sim N}$, we employ a retrieval module $\bm{R}$ to perform text-to-3D searches within an aligned text-3D database \jr{$\bm{B}$}. 
In addition to the mesh structures, each object in the database has been transformed into the 3D representation and also includes bounding box %data
\jr{annotations}, denoted as $\bm{bbox}_{1 \sim M} \in \bm{B}$\jr{, where $\bm{M}$ is a far larger number compared with typical object quantity $\bm{Q}_n$ in an user's normal request.} The %inherent
\jr{retrieved} 3D information %about
\jr{$\bm{bbox}_{1 \sim N}$ of all requested $\bm{N}$} %each 
objects %helps us embed it into the input of the fine-tuned LLM, serving 
\jr{serves} as \jr{the core of} %a
conditional prompt for the LLplace. 
The LLM is tasked with generating the central coordinate $\bm{c}_{1 \sim N}$, and rotation angle data $\bm{r}_{1 \sim N}$.
%Additionally, each generation requires the inclusion of appropriate instruction prompts $\bm{ins}$ to assist the LLM in the generation process.
\jr{Besides, we annotate the meta prompt template of static generation as $\bm{P}_{gen}$, which is a fixed text wrapper of sorting retrieved $\bm{bbox}_{1 \sim N}$ into a fluent text instruction, and then guiding the LLM for effective design. The required input and output of LLplace can be denoted as follows:}
%Thus, the required output of an object from the model is as follows:
% $$
% \{\mathcal{I}, [c, r]_n\} = LLM (ins, (\mathcal{I}, bbox_n)).
% $$
\begin{equation}
  \label{eqa:re}
  \bm{bbox}_{1 \sim N} = \bm{R}(\bm{D}_{1 \sim N}, \bm{B})
\end{equation}
\begin{equation}
  \label{eqa:llm}
  \{\mathcal{I}, [\bm{c}_n, \bm{r}_n]_{1 \sim N}\} = LLM[\bm{P}_{gen}(\mathcal{I}, \bm{bbox}_{1 \sim N})]
\end{equation}
For a single room, there are \jr{$\bm{N}$} objects \jr{to be} placed, which \jr{is} formulated as $\{\bm{bbox}_n, \bm{c}_n, \bm{r}_n\}\in Room$. 
Each triplet $\{\bm{bbox}_n, \bm{c}_n, \bm{r}_n\}$ uniquely determines \jr{the position of $\bm{n}$-th object} in 3D space. \jr{Particularly}, $\bm{bbox}_n = \{\bm{h}_n,\bm{w}_n,\bm{d}_n\}$, which represents the %3D object
bounding box size with height, width, and depth. And $\bm{c}_n = \{\bm{x}_n, \bm{y}_n, \bm{z}_n\}$, which \jr{indicates} the 3D coordinate center and represents the \jr{centroid coordinates}.  
And $\bm{r}_n$ is the rotation angle% of the object
. \jr{Moreover, as previously mentioned, our LLplace support further scene editing. Let $\bm{E}$ denotes either adding ($\bm{+}$) or removing ($\bm{-}$) operation, and $\mathcal{E} = \left\{\bm{E}, \left[ \bm{Q}_k, \bm{D}_k \right]_{1 \sim K} \right\}$ denotes the new editing request over $\bm{K}$ targeting objects:
\begin{equation}
  \label{eqa:add}
  \bm{bbox}_{1 \sim K} = \bm{R}(\bm{D}_{1 \sim K}, \bm{B}|\bm{E}:=\bm{+})
\end{equation}
\begin{equation}
  \label{eqa:edit}
  \{\mathcal{I} \pm \mathcal{E}, [\bm{c}_{n \pm k}, \bm{r}_{n \pm k}]_{1 \sim N \pm K}\} = LLM[\bm{P}_{edit}(\{\mathcal{I}, [\bm{c}_n, \bm{r}_n]_{1 \sim N}\}, \mathcal{E}, \bm{bbox}_{1 \sim K | E:=+})]
\end{equation}
If the $\bm{k}$-th object is to be removed, the user is allowed to provide a new description $\bm{D}_k$ and the removing request $\bm{E}:=\bm{-}$. To add objects, newly bounding box $\bm{bbox}_{1 \sim K}$ will be additionally retrieved. The LLM then interactively modifies the existing layout, using the meta prompt template of editing $\bm{P}_{edit}$ as the text wrapper similar to the static generation stage.} Subsequently, \jr{we parse the output $\{\mathcal{I}, [\bm{c}_n, \bm{r}_n]_{1 \sim N}\}$ of static generation and the output $\{\mathcal{I} \pm \mathcal{E}, [\bm{c}_{n \pm k}, \bm{r}_{n \pm k}]_{1 \sim N \pm K}\}$ of dynamic editing into layouts, and then render them into 3D representations $\bm{S}_{3D}^{gen}$ and $\bm{S}_{3D}^{edit}$, respectively.}

\subsection{LLplace}\label{llplace}
\jr{In this section, we present details of building LLplace.} We employ the \jr{SOTA} open-source model Llama3\footnote{\url{https://huggingface.co/meta-llama/Meta-Llama-3-8B-Instruct}} as a prior foundation. Building on this, we utilize Low-Rank Adaptation (LoRA) for fine-tuning, aiming to stimulate and refine the model’s abilities in 3D world spatial reasoning and configuration \jr{parameter-effectively}.
%This strategy leverages the inherent strengths of Llama3 in handling complex tasks.
%To further adapt the fine-tuning, 
We propose the following %strategies to enhance
\jr{steps to foster} the LLM's understanding, generating and editing of 3D spaces: (1) \textbf{Define Input and Output Format.} Defining the input and output stream of the LLM is a primary task, as appropriate inputs and outputs can significantly aid the model in more effectively generating and understanding 3D spatial features; (2) \jr{\textbf{Establish Meta Prompt Template for Object Placement.} Design instructive meta prompt templates as language wrapper to guide and give reasonable constraints to the LLM for appropriate object placements; (3) \textbf{Construct Dialogue-based Training Data.} Modify and augment the existing indoor scene layout dataset to enhance the LLM's understanding of spatial positions, and leverage its chatting talents for further editing functionality of objects through conversational instructions.}
% \begin{itemize}
%     \item \textbf{Define input and output Format.} Defining the input and output %parameters
%     \jr{stream} of the LLM is a primary task, as appropriate inputs and outputs can significantly aid the model in more effectively generating and understanding 3D spatial features.
%     \item \textbf{Instruction Prompt for Object Placement.} Design precise and accurate prompt templates to guide and give reasonable constraints to the LLM for appropriate object placements. 
%     \item \textbf{Dialogue-based Training Data.} Modify and augment the existing indoor scene layout dataset to enhance the large language model's understanding of spatial positions, while also providing add/remove functionality for objects within the scene as dialogue-based data.
% \end{itemize}

\subsubsection{Define Input and Output Format} \label{inputoutput}
We begin by discussing the input and output specifications of the LLM model during the layout generation \jr{and scene editing} process.
As mentioned in Section~\ref{formulation}, \jr{we can use} the spatial feature %$f_{3D}$ 
\jr{triplet $\{\bm{bbox}, \bm{c}, \bm{r}\}$ to represent an object in a 3D space, where bounding box $\bm{bbox} = \{\bm{h}, \bm{w}, \bm{d}\}$, location coordinates $\bm{c} = \{\bm{x}, \bm{y}, \bm{z}\}$, in specific. In order to provide a cornerstone for the LLM, w}hen we retrieve an appropriate 3D object, %based on the user's text input, 
we determine the %part $f_{3D}$ where $bbox = \{h,w,d\}$ 
\jr{$\bm{bbox}$} as an intrinsic feature that is already inherent in each 3D object, whereas $\bm{c}$ and $\bm{r}$ represent two \jr{other} properties that can be %modified
\jr{freely designed} in space. 
Therefore, in %our model
\jr{LLplace}, we %prefer to leverage 
\jr{merge} the $\bm{bbox}$ feature as the conditional information %for
\jr{with user's description $\bm{D}$, wrapped through carefully designed meta prompt templates $\bm{P}$, }%the prompt input, along with the user’s textual instructions
to feed into the LLM %to assist 
\jr{for} the inference of the spatial information $\bm{c}$ and $\bm{r}$. \jr{The left upper corner of Figure \ref{main_pipeline} illustrates the text-3D retrieval and instruction packing, resulting in the complete LLM's instruction within the \textbf{\color{orange}{orange}} box at the middle upper part, including the step-by-step \textbf{task description} and formalized \textbf{user inputs}. We standardize the user textual inputs with a series of special delimiters and formatted JSON. The room type $\bm{T}$ is wrapped with \textbf{[Task Room Type]} and \textbf{[/Task Room Type]}. And the descriptions of requested objects $[\bm{D}_n, \bm{Q}_n]_{1 \sim N}$ and retrieved $\bm{bbox}_{1 \sim N}$ is correspondingly placed in JSON, attached with \textbf{[Task Objects \& Bounding Box Size]} and \textbf{[/Task Objects \& Bounding Box Size]}.} We aim for the model to infer from the \jr{user's request} and generate the 3D coordinates $\bm{c}_{1 \sim N}$ and rotation angles $\bm{r}_{1 \sim N}$ of objects in the room space. \jr{As demonstrated in the right upper of Figure \ref{main_pipeline}, we guide the model to also report design plans with JSON format, sorting given priors and inferred attributes as key-value pairs, and suggesting with special delimiters \textbf{[Task Output]} and \textbf{[/Task Output]} for easy termination. The ``JSON-in'' and ``JSON-out'' schema enhances the stability of following transcription from the text output into 3D layout (left bottom corner of Figure \ref{main_pipeline}).}
%, denoted as \( c = \{x, y, z\} \), and the objects' rotational angles in space, denoted as \( r \). We will also format the output as JSON and define special tokens for the output between \textbf{[Task Output]} and \textbf{[/Task Output]}.

%For the user textual inputs, we set \jr{a series of} special start delimiter \textbf{[Task Room Type]} and end delimiter \textbf{[/Task Room Type]}.  
%\jr{For the $\bm{n}$-th object, we combine its number $\bm{Q}_n$ and the description $\bm{D}_n$.}
%For example, if the description of a nightstand is `a nightstand with three drawers' and the number of objects is 2, the $[obj]$ should be `2 nightstands with three drawers'.
%Based on the textual description, we retrieve the corresponding object and its bounding box information. 
%To enable the LLM to better understand the description of the object and its bounding box information, we structure the data in JSON format (as shown in Figure 1 input part).
%Additionally, we introduce another special token, \textbf{[Task Objects \& Bounding Box Size]} and \textbf{[/Task Objects \& Bounding Box Size]}, to standardize the specific placement of input textual instruction data.

After generating the indoor scene layout, if the resulting scene is unsatisfactory, the model's dynamic editing capabilities allow for further modifications. 
The input and output text for layout editing are similar to those used in the generation process, \jr{as shown in the bottom of Figure \ref{main_pipeline}}. 
To add an object $\bm{k}$ to the indoor scene, an additional object description $\bm{D}_k$ is used to retrieve the appropriate 3D object and its bounding box information $\bm{bbox}_k$. \jr{The above information, is then formatted into JSON and enclosed within the special delimiters \textbf{[Add Objects]} and \textbf{[/Add Objects]}.}
%An appropriate description template is then applied, facilitating the generation of the complete layout with the newly added object between \textbf{[Added Output]} and \textbf{[/Added Output]}.
\jr{On the contrary}, for deleting objects, the designer does not need to search for and describe the %corresponding 
\jr{exact} 3D object features. We allow users to describe the objects %they see 
using \jr{plain} natural language%without needing to provide a complete description of the objects
. We then convert these descriptions $\bm{D}_{1 \sim K}$ into the format required for removing the objects and \jr{also} 
placing them within special task indicating delimiters \jr{\textbf{[Delete Objects]} and \textbf{[/Delete Objects]}.}

\subsubsection{Establish Meta Prompt Template for Object Placement.} \label{template}
%Further research is needed on how to enhance the capability of large language models to generate spatial layouts of the physical world.
Incorporating well-defined prompt instructions into the methodology of LLMs profoundly influences their capacity to generate and understand spatial layouts. \jr{Except for the aforementioned ``JSON-in'' and ``JSON-out'' schema and tailored component delimiters, we} propose several key %prompt 
\jr{task} statements to guide the generation of layouts. Here's how these guidelines are structured, as briefly demonstrated in the middle top of Figure \ref{main_pipeline} and completely reported in the appendix \ref{prompt template}:
%By explicitly defining the LLM's role as a layout designer, we can establish specific spatial constraints for design. 
%These prompts act as structured guides that shape the model’s focus toward critical spatial relationships and elements within specific scenarios. 
%By utilizing precise prompts and corresponding rules, the approach explicitly directs the synthesis and reconstruction processes of the models, enhancing their ability to produce more accurate and lifelike 3D scenes.

%To further enhance the understanding of 3D spatial features by large language models (LLMs), we propose several key prompt statements to guide the generation of layouts. By explicitly defining the LLM's role as a layout designer, we can establish specific spatial constraints for design. 
\begin{itemize}
    \item \textbf{(1) Placement at Room Edges}: The most important constraint we introduce at first is to encourage the placement of objects at the edges of the room whenever possible. This not only prevents the layout from being too concentrated in the center of the room but also enhances the perception of space, making the room appear larger and more open. This strategy is vital for maximizing the utility of space while maintaining an aesthetically pleasing arrangement, leading to human-preferred design.
    \item \textbf{(2) Avoiding Overlap in Bounding Boxes}: It is crucial to explicitly state in the prompts that the bounding boxes of the generated objects should not overlap. This ensures that the generated layout maintains both functionality and visual appeal. By preventing physical interference between objects, we enhance the usability and aesthetic quality of the space.
    \item \textbf{(3) Alignment of Objects}: To maintain order and symmetry in the layout, the prompt should encourage the alignment of objects. This alignment is essential for aesthetic consistency and functional design, contributing to a harmonious and efficient environment.
    \item \textbf{(4) Setting the Center of 3D Space}: To anchor the model's understanding of space, we define the center of the room as coordinates (0, 0, 0). %We also specify the output format using a JSON template to standardize how layout details are communicated. This template includes keys for object names, their coordinates, and rotation angles, ensuring structured and predictable output. 
\end{itemize}
Additionally, to familiarize the model with %this format 
\jr{JSON inputs and outputs} without biasing its generative process, we provide an \jr{in-context} example that illustrates the format for layout generation. \jr{It is worth noting that this embedded example is a fixed format illustration, which is} not retrieved through any existing \jr{large scale} layout \jr{database (e.g., LayoutGPT)}. 
%but is crafted to exemplify the expected format, fostering a better understanding of input-output handling in the model.
To enable the model to perform further layout edits, we incorporate additional %prompts
\jr{task statements} that guide the editing process% using the LLM
.
Complete prompt templates for generation and editing, which facilitate targeted adjustments and enhancements to the generated layouts, are detailed extensively in the appendix~\ref{prompt template}.
%And to further enhance the training of the LLplace Designer, we tend to build a dialogue-based indoor scene dataset based on the existing dataset.
%These prompt guidelines are designed to help the LLM produce more realistic and practical 3D spatial layouts by embedding specific spatial understanding tasks directly into the model's training framework. This method leverages the LLM’s capabilities in pattern recognition and generation, focusing its output on meeting specific design criteria crucial for real-world applications.

\subsubsection{Construct Dialogue-based Training Data} \label{dialogue}
We apply the 3D-Front~\cite{3d-front} dataset %as
\jr{following} the previous works InstructScene, DiffuScene, and LayoutGPT~\cite{lin2024instructscene,tang2023diffuscene,feng2024layoutgpt}. \jr{We tend to reconstruct the 3D-Front dataset into two turns of dialogue, involving a first turn of static generation and a second turn of dynamic editing. We fully leverage the defined input and out format in section \ref{inputoutput} and established meta prompt templates in section \ref{template}
.}
%As mentioned in LayoutGPT~\cite {feng2024layoutgpt}, since the data of the `living room' is only the few training data, using it directly for supervised training could easily lead to overfitting issues.

\jr{We report our dataset construction algorithm in Alg.~\ref{alg:scene_editing}. In specific, we begin with extracted objects $[\bm{Q}_n, \bm{D}_n]_{1 \sim N}$ according to the design request, along with their attributes, shown as line 5 to 9. 
We then pre-generate complete design input $\bm{input}$ and label $\bm{label}$ with line 10 and 11. 
Afterward, we incorporate randomness for editing functionality, and prepare a corrupted subset ($\bm{input}_{N-K}$, $\bm{label}_{N-K}$) using line 12 to 16. 
For creating addition editing data, we use the subset input $\bm{input}_{N-K}$ and addition objects $\bm{obj}_{1 \sim K}$. 
As for removing data, we use the complete set $\bm{input}$ and removing objects $\bm{obj}_{1 \sim K}$. 
Simultaneously, we comprehensively modify the inputs \& outputs of both generation (the first turn) and editing (the second turn) to contribute a fluent two-round dialogue serving as our training set, as shown from line 17 to 25.} 
Particularly, the \textbf{<|eot\_id |>} is the end of the turn token of the Llama3 model, which means the end of one turn conversation.
Although dialogue data is quite common in LLM tasks, we are the pioneers in creating a dialogue dataset specifically for indoor scene design. 
By engaging in realistic dialogue scenarios, the model can learn to respond dynamically to user requests, simulating real-world interior design consultations.

%We aim to expand the existing LLM's capabilities in understanding 3D spaces, not just limit it to generating fixed LLM scene outputs.
%To enrich the model's capabilities, making it more adept at handling complex spatial placement tasks and 3D spatial understanding in a conversational context, we expand the dialogue-based data for indoor designing.
%From the original data provided in InstructScene~\cite{lin2024instructscene}, each sample denoted as $OD$, the generating and editing data construction process is shown in Alg.~\ref{alg}. 

Finally, each single data sample is annotated with the names of objects in the room, their corresponding 3D data, spatial coordinates, the size of each object's bounding box, and their rotation angles. \jr{The descriptions of the objects are written by GPT-4V and cross-validated by human experts.}

\begin{algorithm*}[!t]
\label{alg}
\caption{Construction of the dialogue data from original 3D-front data.}\label{alg:scene_editing}
\begin{algorithmic}[1]
\REQUIRE The 3D-front data $\bm{OD}$,generation prompt $\bm{P}_{gen}$, and editing prompt $\bm{P}_{edit}$
\ENSURE Construct the dialogue-based generation and editing data with add and remove operations.
\STATE Define the room type $\bm{T}$, objects quantity $\bm{Q}$ and objects description $\bm{D}$
\STATE Define empty sets $\bm{L}_{gen}$ and $\bm{L}_{edit}$ for collecting generation data and editing data, respectively
\FOR{each original data in $\bm{OD}$}
    \STATE Extract design request $\mathcal{I} = \left\{ \bm{T}, \left[ \bm{Q}_n, \bm{D}_n \right]_{1 \sim N} \right\}$
    \FOR{object in $[\bm{Q}_n, \bm{D}_n]_{1 \sim N}$}
        \STATE $\bm{obj} = [\bm{Q}_n, \bm{D}_n]$ %\COMMENT{test}
        \STATE Extract each object's bounding box $\bm{bbox}_n$
        \STATE Extract each object's placement coordinates $\bm{c}_n$ and rotation angle $\bm{r}_n$
    \ENDFOR
    %\STATE <------We build generation data as the first turn of dialogue------>
    \STATE Generate the $\bm{input} = \bm{P}_{gen}[\mathcal{I},\bm{bbox}_{1 \sim N}] + \textbf{<|eot\_id|>}$
    \STATE Generate the $\bm{label} = [\mathcal{I},[\bm{c}_n,\bm{r}_n]_{1 \sim N}]$
    %\STATE <------We build editing data as the second turn of dialogue------>
    \IF{$n > 4$}
        \STATE Randomly select editable objects $\bm{obj}_{1 \sim K} = [\bm{Q}_{k}, \bm{D}_{k}]_{1 \sim K}$ from $\bm{input}$ with probability 0.4
        \STATE Ensure essential items (\textit{table, chair, sofa, bed, lamp}) are not selected
        \STATE Remove $\bm{obj}_{1 \sim K}$ and its attributes from $\bm{input}$ and $\bm{label}$ to get $\bm{input}_{N-K}$ and $\bm{label}_{N-K}$
    \ENDIF
    \IF{Generate editing data for adding objects}
        \STATE Generate the $\bm{input}_{edit} = \bm{P}_{edit}(\bm{label}_{N-K}, [\bm{obj}, \bm{bbox}]_{1 \sim K} | add=True) + \textbf{<|eot\_id|>}$
        \STATE Generate the $\bm{label}_{edit} = \bm{label}$
        \STATE Update the $\bm{input} := \bm{input}_{N-K}$ and the $\bm{label} := \bm{label}_{N-K}$
        % \STATE Update $\bm{input}_{add}$ with add prompt $\bm{P}_{add}$ and add object list $\bm{add\_obj} = \bm{temp[obj,bbox]}$, $\bm{input}_{add} = \{\bm{P}_{add},\bm{add\_obj}\}$
        % \STATE Update $\bm{output}_{add}$ = $\bm{output}$
        % \STATE Update $\bm{P}_{add} = \{\bm{P},\bm{input}_{temp},\bm{output}_{temp}\}+ \bm{<|eot\_id |>}$
        % \STATE Get addition dialogue data sample $\bm{s}_{add} = \{\bm{P}_{add}, \bm{input}_{add}, \bm{output}_{add} \}$
    \ENDIF
    \IF{Generate editing data for removing objects}
        \STATE Generate the $\bm{input}_{edit} = \bm{P}_{edit}(\bm{label}, [\bm{obj}, \bm{bbox}]_{1 \sim K} | remove=True) + \textbf{<|eot\_id|>}$
        \STATE Generate the $\bm{label}_{edit} = \bm{label}_{N-K}$
        % \STATE Update $\bm{input}_{re}$ with remove prompt $\bm{P}_{re}$ and remove object list $\bm{remove\_obj} =\bm{temp[obj]}$, $\bm{input}_{re} = \{\bm{P}_{re},\bm{remove\_obj}\}$
        % \STATE Update $\bm{output}_{re}$ = $\bm{output}_{temp}$
        % \STATE Update $\bm{P}_{re} = \{\bm{P},\bm{input},\bm{output}\}+ \bm{<|eot\_id |>}$
        % \STATE Get removal dialogue data sample $\bm{s}_{re} = \{\bm{P}_{re}, \bm{input}_{re}, \bm{output}_{re}\}$
    \ENDIF
    \STATE Add ($\bm{input}$, $\bm{label}$) data pair to generation set $\bm{L}_{gen}$
    \STATE Add ($\bm{input}_{edit}$, $\bm{label}_{edit}$) data pair to generation set $\bm{L}_{edit}$
\ENDFOR
%\STATE Extract $[\bm{obj}, \bm{c}, \bm{r}]_n$ from $\bm{output}$ of $\bm{s}$

\end{algorithmic}
\end{algorithm*}

%This method ensures that the model can dynamically edit scenes by adding or removing objects based on structured dialogue inputs, thereby enhancing its capability to understand and manipulate 3D space."

%%how to organize the input is another problem. 

\section{Experiments}

\subsection{Experiments Setup}
\noindent\textbf{Dataset Setup.}
%After data construction, 
Our constructed dataset contains a total of 5,754 original entries. Through the process of random removal and addition%and generating dialogue templates
, we create an additional 2,944 removal editing instructions and 3,100 addition editing instructions.
We use the same room test set as ATISS~\cite{paschalidou2021atiss} and LayoutGPT~\cite{feng2024layoutgpt}, consisting of 423 bedroom entries and 53 living room entries. We also use 43 bedroom editing entries and 33 living room editing entries in the test dataset. We split the rest of the 11,246 data with 566 as the eval set and other data as the training set. 

\begin{table*}[!t]
    \centering
    \caption{The quantitative performance between LLplace and LayoutGPT.}
    \resizebox{0.9\linewidth}{!}{
    \begin{tabular}{clcc|ccc}
    \toprule
    Method & Room Type & FID ↓ & OOR ↓ &GPT-4o Func. ↑&GPT-4o Layout. ↑&GPT-4o Aes. ↑ \\
    \midrule
    \multirow{3}{*}{ LayoutGPT~\cite{feng2024layoutgpt}} & Bedroom & 133.7  & 0.078&8.0 &7.5&7.2 \\
     & Livingroom & 171.6  &0.487& 7.2 & \textbf{7.8} & 7.2\\
     & Avg. & 152.7 &0.283& 7.6&7.65&7.2 \\
    \midrule
    \multirow{3}{*}{ LLplace} & Bedroom & \textbf{95.8}  & \textbf{0.056}&\textbf{8.4}&\textbf{8.0}&\textbf{7.5}  \\
     & Livingroom & \textbf{157.8}  &\textbf{0.472}&\textbf{7.6}&7.6&7.2 \\
     & Avg. & \textbf{126.8} &\textbf{0.264}&\textbf{8.0}&\textbf{7.8}&\textbf{7.35}  \\
    \bottomrule
    \end{tabular}}
    \label{weightbalance}
    \vspace{-1em}
\end{table*}

\begin{figure*}[!t]
	\centering
	\includegraphics[width=1\linewidth]{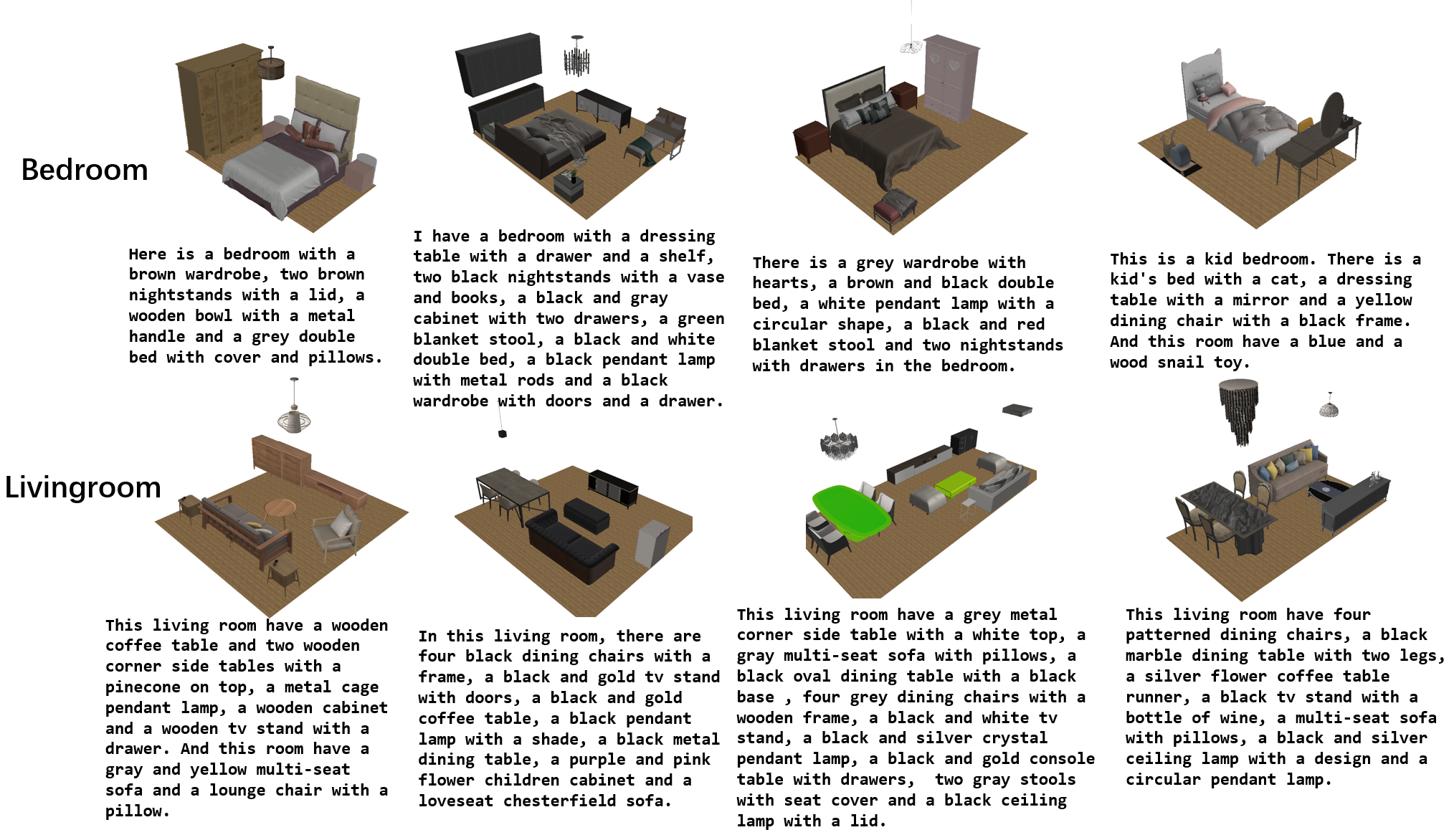}
 \vspace{-2em}
	\caption{The showcases of LLplace in generating the layout of the bedroom and the living room.
}
\label{results}

\end{figure*}

\noindent\textbf{Training Setup.}
In our training process, due to resource constraints, we adopt the latest open-source model Llama3-8B-Instruct as the base LLM and use LoRA~\cite{hu2021lora} to fine-tune Llama3, for training the LLM's 3D layout generation and editing understanding capabilities. Specifically, we set the LoRA alpha value to 32, the LoRA r value to 8, and the LoRA dropout rate to 0.05. Additionally, we set the learning rate to 1e-4 and use a cosine scheduler for optimization, training for 20 epochs.
All the experiments are conducted on four NVIDIA A100 X 40G GPUs. It takes 18 hours for entire training.

\noindent\textbf{Evaluation Metrics.}
Existing methods do not provide a completely unified set of evaluation metrics. Therefore, in this paper, we introduce three evaluation metrics. The first is the FID metric, as used in LayoutGPT~\cite{feng2024layoutgpt}, to assess the consistency between the rendered and real scenes. 
Next, we calculate the Object Overlap Rate (OOR) of bounding boxes in the scene to evaluate the rationality of the generated scene layout. 
Finally, we use the GPT-4o model to assess \jr{the quality of} our rendered results. 
We use a prompt template to make GPT-4o evaluate the generated layouts from three perspectives: \textbf{Functionality and Activity-based Alignment}, \textbf{Layout and Furniture}, and \textbf{Aesthetics of the Room's Layout}. \jr{Each aspect has a marking score varying from 0 to 10.}
%maximum score of 10 points. 
%Although the results obtained using GPT-4o for evaluation are not always consistent, it provides an objective method for assessing the feasibility of the generated layouts.
To maintain consistency in rendering, we apply the open-source simple 3D viz renderer for all renderings.

\subsection{Experiment Results}
\noindent\textbf{Quantitative Results.}
From the OOR values, it is evident that our bounding box position predictions yield a more reasonable distribution. Despite the OOR values indicating that the layout generation quality for living rooms still lags behind that of bedrooms, compared with LayoutGPT, which requires expensive high-quality in-context exemplars, we lead 2.2\% and 1.5\% OOR values in the bedroom and living room scenarios, respectively.
The FID values further demonstrate that our method can produce higher-quality scenes. LLplace achieves the best generation results for bedrooms, and for living rooms, obtaining up to 25.9 absolute improvements on average, compared with LayoutGPT.
\jr{Furthermore}, the evaluation results using GPT-4o show that our model performs better in rendered scenes for Bedrooms and has a similar performance to LayoutGPT in living Rooms, \jr{across all three evaluation aspects}.
In terms of functionality, GPT-4o considers our generated room layouts to be more practical\jr{, marking with a higher 8.0 average score}. In terms of layout rationality and aesthetics, our model still leads LayoutGPT with \jr{0.15 absolute improvements for both perspectives.}
%a slight performance advantage.
%However, this metric can only assess the overall rendering result and has limitations, such as its inability to accurately reflect the rotation angles of each object in the environment.

\begin{figure*}[!t]
	\centering
	\includegraphics[width=0.85\linewidth]{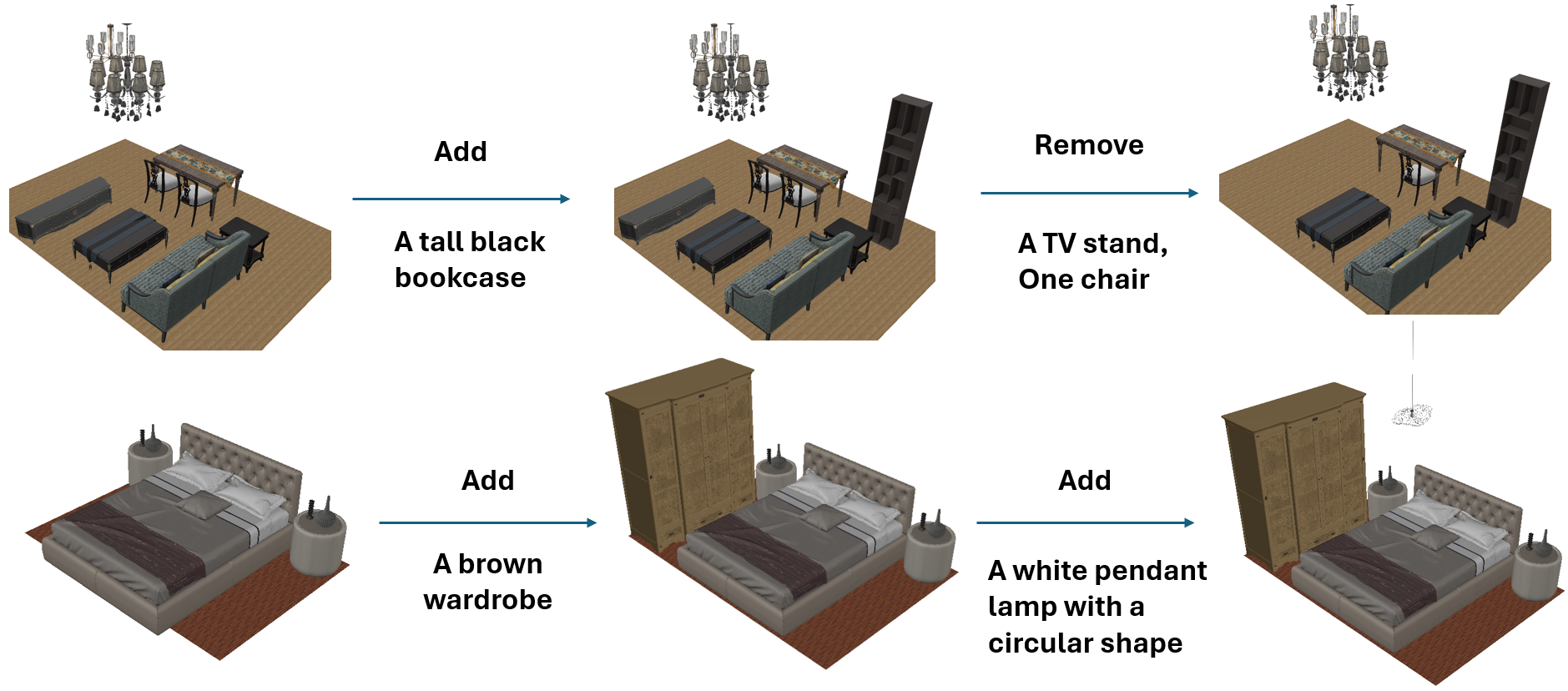}
	\caption{The qualitative results of the LLplace in scene editing.
}
\label{editing}
\end{figure*}

\noindent\textbf{Qualitative Reports.}
As shown in Fig.~\ref{results}, the LLplace can generate reasonable layouts and understand the general 3D relationship between objects of the indoor scene.
For example, \jr{Llplace accurately understand the following spatial relations:} ``\emph{the wardrobe should be on the side of the bed}'',  ``\emph{chairs should be placed around the table}'', and ``\emph{the TV stand should be placed in front of the bed or the couch}''.
%Our results are rendered by the open-source renderer simple 3D viz.
\jr{We also provide further comparison analysis of cases presented in Figure \ref{first image} in Appendix~\ref{more_results_appendix}, across our LLplace and other strong baselines.}
%Additionally, we compared our method with the results of LayoutGPT and used \jr{our meta prompt template $\bm{P}_{gen}$ to obtain the 3D layout designed by GPT-4o}. 
%More comparison results with LayoutGPT and GPT-4o with our meta prompt are shown in Appendix~\ref{more_results_appendix}.

\noindent\textbf{Scene Editing Results.}
In Figure~\ref{editing}, we demonstrate the scene editing capabilities of LLplace. As shown in the figure, we can add objects to an existing scene through language instructions. For example, we added a tall bookshelf to the living room. LLplace can analyze the relationships between existing objects and place the additional object in a suitable position. If the user wants to delete an object, they can do so using simple descriptive keywords \jr{as well}. For instance, ``\emph{a TV stand}'' and ``\emph{one chair}'' are removed from the scene \jr{at the first row}. This proves that LLplace, trained with dialogue data, possesses scene understanding and editing capabilities that current models cannot achieve. 
We also test the FID and OOR %results 
\jr{scores of editing} %with 
random 20 %additional 
bedroom scenes and 20 living room scenes. As shown in Table~\ref{editing table}, %the FID and OOR have stable performance as the generated scenes.
\jr{LLplace maintains a relatively stable performance even after editing, losing 16.6 FID average score, which is still better than the 152.7 FID score of LayoutGPT in Table \ref{weightbalance}.}

\subsection{Ablation Study}
We apply random 50 bedroom scenes to conduct the ablation study.
We test the generation results without adding key task instruction introduced in section \ref{template} and evaluate the object addition capability of the model trained only with the original data, excluding dialogue data. 
We also report the FID and OOR values
% to demonstrate the importance of task instruction and dialogue data
. As shown in the upper half of Table~\ref{ablation}, %even without using task instruction, the trained model can still achieve good generation capabilities. However, there is a significant gap compared to using task instruction. 
\jr{the removal of task instruction leads to a significant performance degradation. Nevertheless, involving our carefully designed ``JSON-in/out'' schema and dialogue data during the training promises a comparable performance compared with former strong baselines.}
And the model trained with dialogue data can better understand the scene and place the added objects in appropriate locations, \jr{as reported in the second row of Table \ref{ablation}}.

\begin{table}[t]
\begin{minipage}[t]{0.49\linewidth}
    % \footnotesize
    \centering
    \caption{The FID ↓ and OOR ↓ performance \\after adding objects in the scenes by LLplace.}
        \label{editing table}
    \resizebox{0.8\linewidth}{!}{
    \begin{tabular}{llcc}
        \toprule
        Method & Room  &FID ↓ &OOR ↓ \\
        \midrule
        \multirow{3}{*}{Generation} & Bedroom&95.8& 0.056   \\
         & Livingroom &157.8& 0.472    \\
         & Avg. &126.8 & 0.264  \\
        \midrule
        \multirow{3}{*}{ Editing} & Bedroom& 114.7& 0.060   \\
         & Livingroom &172.4& 0.525    \\
         & Avg. &143.4&  0.293   \\
        \bottomrule
    \end{tabular}}
\end{minipage}
\hfill
\begin{minipage}[t]{0.49\linewidth}
    % \footnotesize
    \centering    
    \caption{The ablation studies of generation without task instruction and editing without training.}
    \label{ablation}
    \resizebox{0.9\linewidth}{!}{
    \begin{tabular}{llcc}
        \toprule
        Method &   &FID ↓ &OOR ↓ \\
        \midrule
        \multirow{2}{*}{Task Ins.(TI.)} & with TI.&95.8& 0.056   \\
         & w/o TI. &131.5 & 0.103  \\
        \midrule
        \multirow{2}{*}{Dialogue Editing} & with training& 114.7& 0.060   \\
         & w/o training &142.1& 0.072    \\
        \bottomrule
    \end{tabular}}
    \end{minipage}
    \vspace{-1em}
\end{table}

\section{Conclusion}
In this paper, we propose LLplace, a novel 3D indoor designer. 
LLplace utilizes LoRA to fine-tune the Llama3-8B-Instruct LLM to enable both room layout generation and scene editing through dialogue. Specifically, our model does not rely on \jr{expensive} in-context exemplars and spatial relationship priors as existing approaches insist.
Instead, we develop general prompt templates, then follow the mainstream paradigm of LLMs to extend the 3D-Front dataset into a dialogue dataset containing one turn of generation and the other turn of further editing, which manages to foster LLplace with both static generation and dynamic editing capabilities. Experimental results demonstrate that LLplace outperforms existing LLM-based indoor scene design methods across various metrics.

\bibliographystyle{plainnat}
\bibliography{neurips_2024}

\begin{thebibliography}{31}
\providecommand{\natexlab}[1]{#1}
\providecommand{\url}[1]{\texttt{#1}}
\expandafter\ifx\csname urlstyle\endcsname\relax
  \providecommand{\doi}[1]{doi: #1}\else
  \providecommand{\doi}{doi: \begingroup \urlstyle{rm}\Url}\fi

\bibitem[Achiam et~al.(2023)Achiam, Adler, Agarwal, Ahmad, Akkaya, Aleman, Almeida, Altenschmidt, Altman, Anadkat, et~al.]{openai2024gpt4}
Josh Achiam, Steven Adler, Sandhini Agarwal, Lama Ahmad, Ilge Akkaya, Florencia~Leoni Aleman, Diogo Almeida, Janko Altenschmidt, Sam Altman, Shyamal Anadkat, et~al.
\newblock Gpt-4 technical report.
\newblock \emph{arXiv preprint arXiv:2303.08774}, 2023.

\bibitem[Aguina-Kang et~al.(2024)Aguina-Kang, Gumin, Han, Morris, Yoo, Ganeshan, Jones, Wei, Fu, and Ritchie]{aguina2024open}
Rio Aguina-Kang, Maxim Gumin, Do~Heon Han, Stewart Morris, Seung~Jean Yoo, Aditya Ganeshan, R~Kenny Jones, Qiuhong~Anna Wei, Kailiang Fu, and Daniel Ritchie.
\newblock Open-universe indoor scene generation using llm program synthesis and uncurated object databases.
\newblock \emph{arXiv preprint arXiv:2403.09675}, 2024.

\bibitem[{\c{C}}elen et~al.(2024){\c{C}}elen, Han, Schindler, Van~Gool, Armeni, Obukhov, and Wang]{ccelen2024design}
Ata {\c{C}}elen, Guo Han, Konrad Schindler, Luc Van~Gool, Iro Armeni, Anton Obukhov, and Xi~Wang.
\newblock I-design: Personalized llm interior designer.
\newblock \emph{arXiv preprint arXiv:2404.02838}, 2024.

\bibitem[Chen et~al.(2023)Chen, Chen, Zhang, Li, Yu, Fei, Zhu, Fan, and Chen]{chen2023ll3da}
Sijin Chen, Xin Chen, Chi Zhang, Mingsheng Li, Gang Yu, Hao Fei, Hongyuan Zhu, Jiayuan Fan, and Tao Chen.
\newblock Ll3da: Visual interactive instruction tuning for omni-3d understanding, reasoning, and planning.
\newblock \emph{arXiv preprint arXiv:2311.18651}, 2023.

\bibitem[Deitke et~al.(2022)Deitke, VanderBilt, Herrasti, Weihs, Ehsani, Salvador, Han, Kolve, Kembhavi, and Mottaghi]{deitke2022️ProcTHOR}
Matt Deitke, Eli VanderBilt, Alvaro Herrasti, Luca Weihs, Kiana Ehsani, Jordi Salvador, Winson Han, Eric Kolve, Aniruddha Kembhavi, and Roozbeh Mottaghi.
\newblock Procthor: Large-scale embodied ai using procedural generation.
\newblock \emph{Advances in Neural Information Processing Systems}, 35:\penalty0 5982--5994, 2022.

\bibitem[Fang et~al.(2023)Fang, Hu, Luo, and Tan]{fang2023ctrl}
Chuan Fang, Xiaotao Hu, Kunming Luo, and Ping Tan.
\newblock Ctrl-room: Controllable text-to-3d room meshes generation with layout constraints.
\newblock \emph{arXiv preprint arXiv:2310.03602}, 2023.

\bibitem[Feng et~al.(2024)Feng, Zhu, Fu, Jampani, Akula, He, Basu, Wang, and Wang]{feng2024layoutgpt}
Weixi Feng, Wanrong Zhu, Tsu-jui Fu, Varun Jampani, Arjun Akula, Xuehai He, Sugato Basu, Xin~Eric Wang, and William~Yang Wang.
\newblock Layoutgpt: Compositional visual planning and generation with large language models.
\newblock \emph{Advances in Neural Information Processing Systems}, 36, 2024.

\bibitem[Fu et~al.(2021{\natexlab{a}})Fu, Cai, Gao, Zhang, Wang, Li, Zeng, Sun, Jia, Zhao, et~al.]{3d-front}
Huan Fu, Bowen Cai, Lin Gao, Ling-Xiao Zhang, Jiaming Wang, Cao Li, Qixun Zeng, Chengyue Sun, Rongfei Jia, Binqiang Zhao, et~al.
\newblock 3d-front: 3d furnished rooms with layouts and semantics.
\newblock In \emph{Proceedings of the IEEE/CVF International Conference on Computer Vision}, pages 10933--10942, 2021{\natexlab{a}}.

\bibitem[Fu et~al.(2021{\natexlab{b}})Fu, Jia, Gao, Gong, Zhao, Maybank, and Tao]{3dfuture}
Huan Fu, Rongfei Jia, Lin Gao, Mingming Gong, Binqiang Zhao, Steve Maybank, and Dacheng Tao.
\newblock 3d-future: 3d furniture shape with texture.
\newblock \emph{International Journal of Computer Vision}, 129:\penalty0 3313--3337, 2021{\natexlab{b}}.

\bibitem[Fu et~al.(2024)Fu, Liu, Chen, Nie, and Xiong]{fu2024scenellm}
Rao Fu, Jingyu Liu, Xilun Chen, Yixin Nie, and Wenhan Xiong.
\newblock Scene-llm: Extending language model for 3d visual understanding and reasoning.
\newblock \emph{arXiv preprint arXiv:2403.11401}, 2024.

\bibitem[Gao et~al.(2023)Gao, Sun, Mo, Lai, Guibas, and Yang]{gao2023scenehgn}
Lin Gao, Jia-Mu Sun, Kaichun Mo, Yu-Kun Lai, Leonidas~J Guibas, and Jie Yang.
\newblock Scenehgn: Hierarchical graph networks for 3d indoor scene generation with fine-grained geometry.
\newblock \emph{IEEE Transactions on Pattern Analysis and Machine Intelligence}, 2023.

\bibitem[Ho et~al.(2020)Ho, Jain, and Abbeel]{ho2020denoising}
Jonathan Ho, Ajay Jain, and Pieter Abbeel.
\newblock Denoising diffusion probabilistic models.
\newblock \emph{Advances in neural information processing systems}, 33:\penalty0 6840--6851, 2020.

\bibitem[Hu et~al.(2021)Hu, Shen, Wallis, Allen-Zhu, Li, Wang, Wang, and Chen]{hu2021lora}
Edward~J Hu, Yelong Shen, Phillip Wallis, Zeyuan Allen-Zhu, Yuanzhi Li, Shean Wang, Lu~Wang, and Weizhu Chen.
\newblock Lora: Low-rank adaptation of large language models.
\newblock \emph{arXiv preprint arXiv:2106.09685}, 2021.

\bibitem[Huang et~al.(2023{\natexlab{a}})Huang, Yong, Ma, Linghu, Li, Wang, Li, Zhu, Jia, and Huang]{huang2023leo}
Jiangyong Huang, Silong Yong, Xiaojian Ma, Xiongkun Linghu, Puhao Li, Yan Wang, Qing Li, Song-Chun Zhu, Baoxiong Jia, and Siyuan Huang.
\newblock An embodied generalist agent in 3d world.
\newblock \emph{arXiv preprint arXiv:2311.12871}, 2023{\natexlab{a}}.

\bibitem[Huang et~al.(2023{\natexlab{b}})Huang, Wang, Li, Jia, Liu, Zhu, Liang, and Zhu]{huang2023diffusionfor3d}
Siyuan Huang, Zan Wang, Puhao Li, Baoxiong Jia, Tengyu Liu, Yixin Zhu, Wei Liang, and Song-Chun Zhu.
\newblock Diffusion-based generation, optimization, and planning in 3d scenes.
\newblock In \emph{Proceedings of the IEEE/CVF Conference on Computer Vision and Pattern Recognition}, pages 16750--16761, 2023{\natexlab{b}}.

\bibitem[Kingma and Welling(2013)]{VAE}
Diederik~P Kingma and Max Welling.
\newblock Auto-encoding variational bayes.
\newblock \emph{arXiv preprint arXiv:1312.6114}, 2013.

\bibitem[Lin and Mu(2024)]{lin2024instructscene}
Chenguo Lin and Yadong Mu.
\newblock Instructscene: Instruction-driven 3d indoor scene synthesis with semantic graph prior.
\newblock \emph{arXiv preprint arXiv:2402.04717}, 2024.

\bibitem[Liu et~al.(2023)Liu, Lei, Wang, Huang, Feng, Wen, Cheng, Ke, Xu, Tam, et~al.]{liu2023alignbench}
Xiao Liu, Xuanyu Lei, Shengyuan Wang, Yue Huang, Zhuoer Feng, Bosi Wen, Jiale Cheng, Pei Ke, Yifan Xu, Weng~Lam Tam, et~al.
\newblock Alignbench: Benchmarking chinese alignment of large language models.
\newblock \emph{arXiv preprint arXiv:2311.18743}, 2023.

\bibitem[Lu et~al.(2023)Lu, An, Lin, Pergola, He, Yin, Sun, and Wu]{lu2023memochat}
Junru Lu, Siyu An, Mingbao Lin, Gabriele Pergola, Yulan He, Di~Yin, Xing Sun, and Yunsheng Wu.
\newblock Memochat: Tuning llms to use memos for consistent long-range open-domain conversation.
\newblock \emph{arXiv preprint arXiv:2308.08239}, 2023.

\bibitem[Luo et~al.(2020)Luo, Zhang, Wu, and Tenenbaum]{luo2020endtoend}
Andrew Luo, Zhoutong Zhang, Jiajun Wu, and Joshua~B Tenenbaum.
\newblock End-to-end optimization of scene layout.
\newblock In \emph{Proceedings of the IEEE/CVF Conference on Computer Vision and Pattern Recognition}, pages 3754--3763, 2020.

\bibitem[Paschalidou et~al.(2021)Paschalidou, Kar, Shugrina, Kreis, Geiger, and Fidler]{paschalidou2021atiss}
Despoina Paschalidou, Amlan Kar, Maria Shugrina, Karsten Kreis, Andreas Geiger, and Sanja Fidler.
\newblock Atiss: Autoregressive transformers for indoor scene synthesis.
\newblock \emph{Advances in Neural Information Processing Systems}, 34:\penalty0 12013--12026, 2021.

\bibitem[Purkait et~al.(2020)Purkait, Zach, and Reid]{purkait2020vae}
Pulak Purkait, Christopher Zach, and Ian Reid.
\newblock Sg-vae: Scene grammar variational autoencoder to generate new indoor scenes.
\newblock In \emph{European Conference on Computer Vision}, pages 155--171. Springer, 2020.

\bibitem[Ritchie et~al.(2019)Ritchie, Wang, and Lin]{ritchie2019fast}
Daniel Ritchie, Kai Wang, and Yu-an Lin.
\newblock Fast and flexible indoor scene synthesis via deep convolutional generative models.
\newblock In \emph{Proceedings of the IEEE/CVF Conference on Computer Vision and Pattern Recognition}, pages 6182--6190, 2019.

\bibitem[Tang et~al.(2023)Tang, Nie, Markhasin, Dai, Thies, and Nie{\ss}ner]{tang2023diffuscene}
Jiapeng Tang, Yinyu Nie, Lev Markhasin, Angela Dai, Justus Thies, and Matthias Nie{\ss}ner.
\newblock Diffuscene: Scene graph denoising diffusion probabilistic model for generative indoor scene synthesis.
\newblock \emph{arXiv preprint arXiv:2303.14207}, 2023.

\bibitem[Touvron et~al.(2023)Touvron, Lavril, Izacard, Martinet, Lachaux, Lacroix, Rozi{\`e}re, Goyal, Hambro, Azhar, et~al.]{touvron2023llama}
Hugo Touvron, Thibaut Lavril, Gautier Izacard, Xavier Martinet, Marie-Anne Lachaux, Timoth{\'e}e Lacroix, Baptiste Rozi{\`e}re, Naman Goyal, Eric Hambro, Faisal Azhar, et~al.
\newblock Llama: Open and efficient foundation language models.
\newblock \emph{arXiv preprint arXiv:2302.13971}, 2023.

\bibitem[Wang et~al.(2021)Wang, Yeshwanth, and Nie{\ss}ner]{wang2021sceneformer}
Xinpeng Wang, Chandan Yeshwanth, and Matthias Nie{\ss}ner.
\newblock Sceneformer: Indoor scene generation with transformers.
\newblock In \emph{2021 International Conference on 3D Vision (3DV)}, pages 106--115. IEEE, 2021.

\bibitem[Yang et~al.(2021{\natexlab{a}})Yang, Zhang, Yan, Huang, Ma, Zheng, Bajaj, and Huang]{yang2021scene_uncentain}
Haitao Yang, Zaiwei Zhang, Siming Yan, Haibin Huang, Chongyang Ma, Yi~Zheng, Chandrajit Bajaj, and Qixing Huang.
\newblock Scene synthesis via uncertainty-driven attribute synchronization.
\newblock In \emph{Proceedings of the IEEE/CVF International Conference on Computer Vision}, pages 5630--5640, 2021{\natexlab{a}}.

\bibitem[Yang et~al.(2021{\natexlab{b}})Yang, Guo, Zhou, and Tong]{yang2021indoor}
Ming-Jia Yang, Yu-Xiao Guo, Bin Zhou, and Xin Tong.
\newblock Indoor scene generation from a collection of semantic-segmented depth images.
\newblock In \emph{Proceedings of the IEEE/CVF International Conference on Computer Vision}, pages 15203--15212, 2021{\natexlab{b}}.

\bibitem[Yang et~al.(2024{\natexlab{a}})Yang, Jia, Zhi, and Huang]{yang2024physcene}
Yandan Yang, Baoxiong Jia, Peiyuan Zhi, and Siyuan Huang.
\newblock Physcene: Physically interactable 3d scene synthesis for embodied ai.
\newblock \emph{arXiv preprint arXiv:2404.09465}, 2024{\natexlab{a}}.

\bibitem[Yang et~al.(2024{\natexlab{b}})Yang, Sun, Weihs, VanderBilt, Herrasti, Han, Wu, Haber, Krishna, Liu, et~al.]{yang2024holodeck}
Yue Yang, Fan-Yun Sun, Luca Weihs, Eli VanderBilt, Alvaro Herrasti, Winson Han, Jiajun Wu, Nick Haber, Ranjay Krishna, Lingjie Liu, et~al.
\newblock Holodeck: Language guided generation of 3d embodied ai environments.
\newblock In \emph{The IEEE/CVF Conference on Computer Vision and Pattern Recognition (CVPR 2024)}, volume~30, pages 20--25. IEEE/CVF, 2024{\natexlab{b}}.

\bibitem[Zhang et~al.(2023)Zhang, Rao, and Agrawala]{zhang2023controlnet}
Lvmin Zhang, Anyi Rao, and Maneesh Agrawala.
\newblock Adding conditional control to text-to-image diffusion models.
\newblock In \emph{Proceedings of the IEEE/CVF International Conference on Computer Vision}, pages 3836--3847, 2023.

\end{thebibliography}

\appendix

\section{Appendix}
\subsection{Details of the Evaluation Metrics}
\noindent\textbf{Object Overlap Rate (OOR).} The object overlap rate (OOR) metric quantifies the spatial intersection between a set of 3D bounding boxes. It is calculated by determining the volume of intersection between each pair of bounding boxes and then dividing this volume by the smaller volume of the two boxes. This metric provides a value indicating the degree of overlap, where a value of 0 indicates no overlap and a value of 1 indicates complete embedding of the smaller box within the larger one. We use the OOR metric to evaluate whether objects within a 3D scene overlap.

\noindent\textbf{GPT-4o Evaluation.} We use an evaluation method similar to that mentioned in I-Design~\cite{ccelen2024design}, where GPT-4o acts as the evaluator, scoring our generated room layouts with a maximum score of 10. The prompt template we used is shown in Table~\ref{gpt evaluation}.

\subsection{Objects Retrieval}
During inference, we chose to perform object retrieval from the 3D-Future dataset~\cite{3dfuture}. We use the item description data annotated by GPT-4V as mentioned in InstuctScene~\cite{lin2024instructscene}, and each retrieval was conducted through text matching. This is because our method does not require the specific features of 3D objects for training or inference; thus, we opt for the most efficient and convenient text-matching retrieval method. Each retrieval yields the file path corresponding to the 3D asset and the bounding box dimensions of the object, allowing us to include it in the input for scene generation or editing.

\subsection{The Prompt Template}
\label{prompt template}
In this section, we present all our prompt templates, including the meta prompt template, the add prompt, and the remove prompt templates.
We show the meta prompt template, the removal prompt template, and the addition prompt template in Table~\ref{meta prompt}, Table~\ref{remove prompt}, and Table~\ref{add prompt}, respectively. 

As shown in Table~\ref{meta prompt}, our complete meta prompt contains a total of 8 rules. Besides the four rules mentioned in the main text, the remaining four further standardize the placement rules for our 3D scene, such as the placement of chairs and the format of the LLM's output. In the task instruction Table~\ref{remove prompt}, \ref{add prompt}, we demonstrate in the table how the instructions for editing should be defined. 

\subsection{More Qualitative Results} \label{more_results_appendix}
\begin{figure*}[!t]
	\centering
	\includegraphics[width=0.85\linewidth]{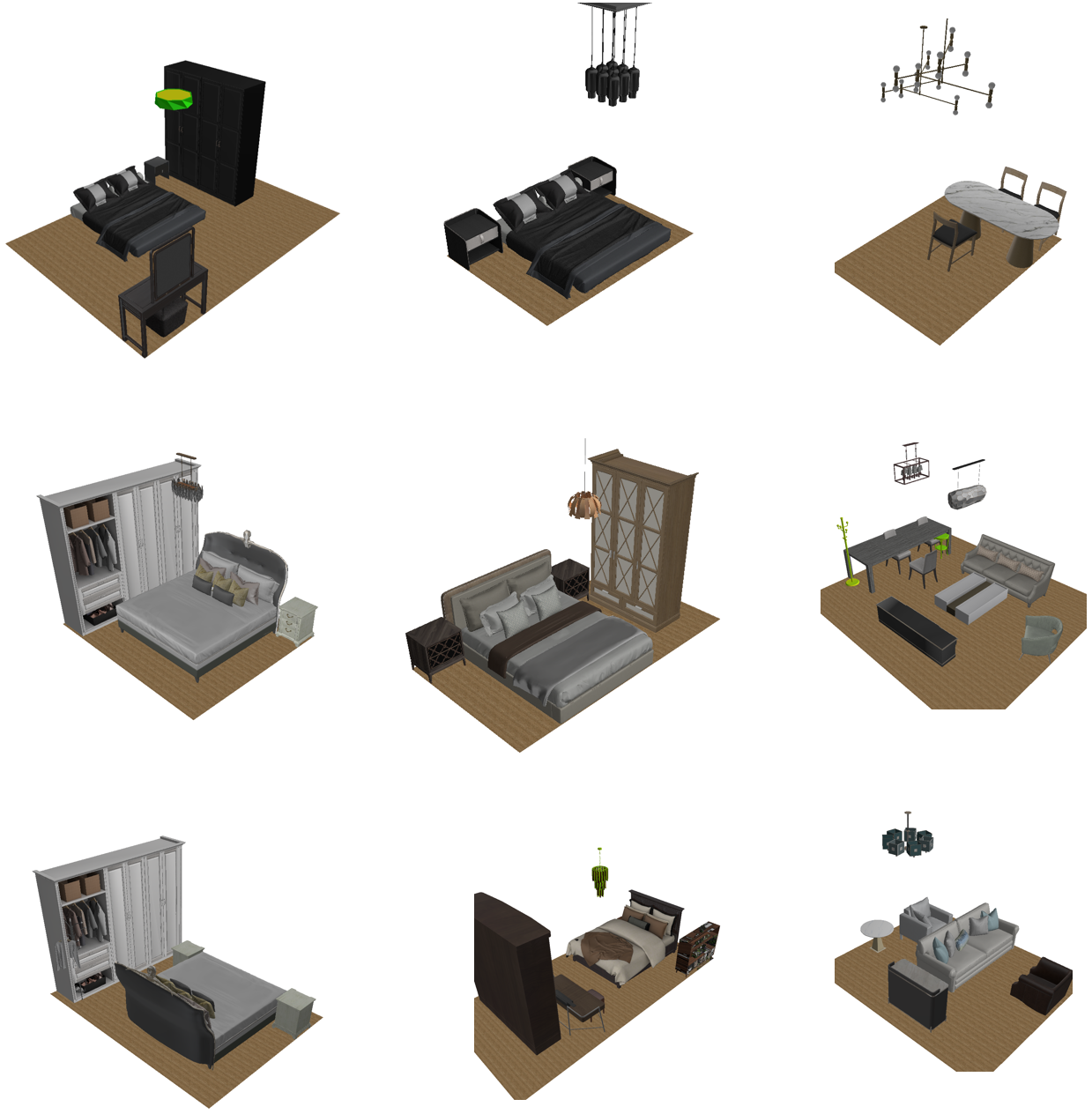}

	\caption{The more showcases of LLplace in generating the 3D scene layouts.
}
\label{moreresults}

\end{figure*}

In Figure~\ref{moreresults}, we illustrate more generated results of the bedroom and living room. These examples demonstrate that our method achieves reasonable results in generating 3D indoor scenes.

\begin{figure*}[!t]
	\centering
	\includegraphics[width=1\linewidth]{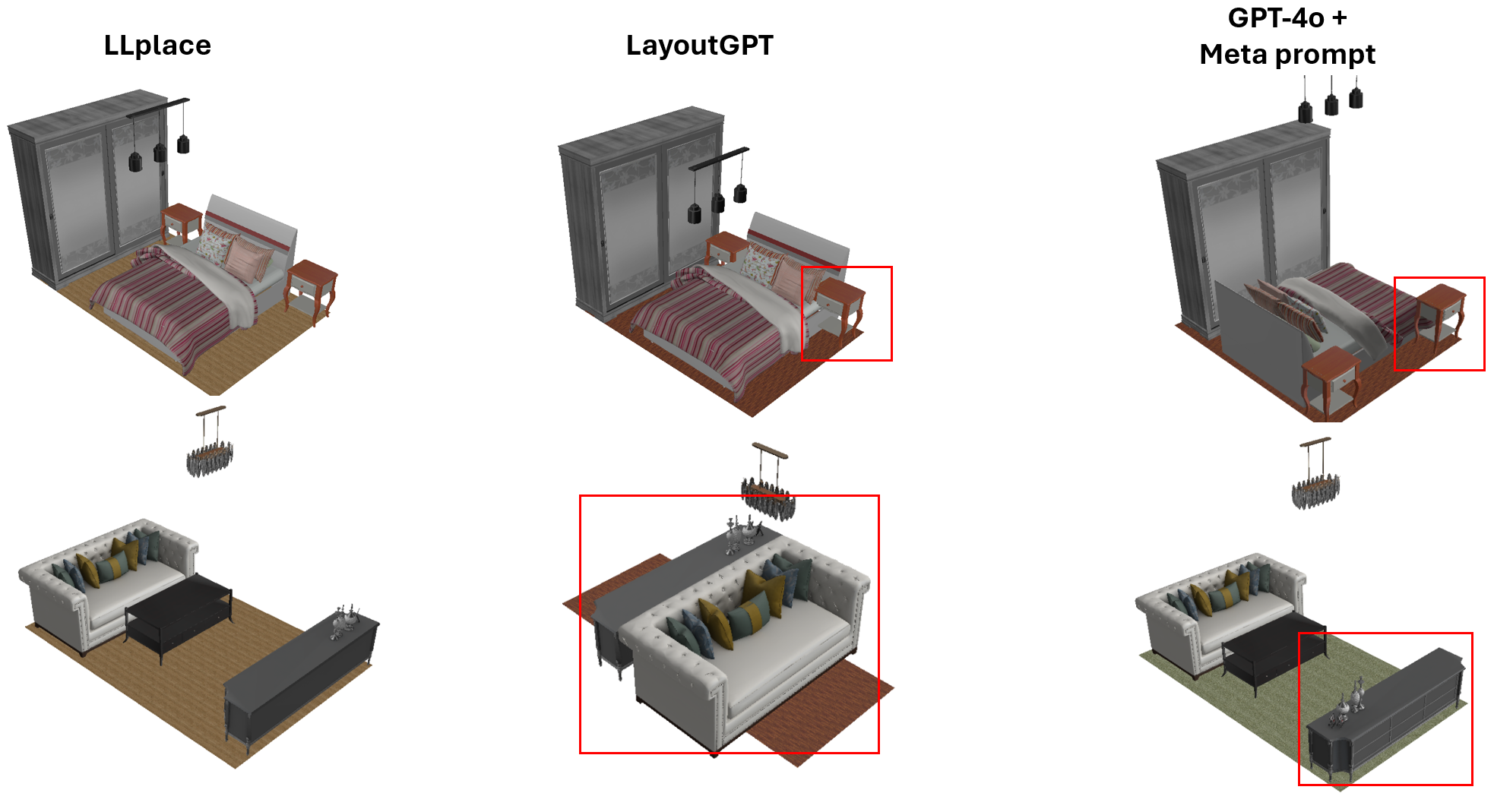}
	\caption{The comparison results of the LLplace, LayoutGPT, and GPT-4o + our meta prompt.
}
\label{compare results}

\end{figure*}

And we compare more generation results with LayoutGPT and GPT-4o+meta prompt in Figure~\ref{compare results}.
As shown in Figure~\ref{compare results}, our model produces a more reasonable distribution of 3D objects and outputs reasonable rotation angles.
LayoutGPT, on the other hand, %relies on in-context exemplars for guidance. This dependence not only leads to 
\jr{fails to produce} occasional overlaps between objects and issue of object rotation angles, which is a drawback of \jr{heavily relying with high-quality examples}.
\jr{Besides}, although GPT-4o does not produce significant overlap issues, it lacks %prior information on 
\jr{vertical adaptation on the} indoor scene \jr{task}, resulting in less plausible room layouts.

\subsection{Limitations}
Although our method extends the LLM-based 3D scene layout generation task to allow for scene editing through dialogue, there are still several limitations that need to be addressed. First, due to resource constraints, we can only fine-tune lightweight large language models using LoRA. While this leverages the existing knowledge base of the LLM, the inability to fully fine-tune means that the LLM's full potential cannot be activated. Second, our model is limited to a token length of 2048, restricting it to once or twice turn scene edits conversation in training and inference. Although we can process the new output to the whole layout JSON for the next turn conversation, we hope to apply more length tokens to train and inference. In the future, we can consider expanding the dataset to construct a larger-scale scene editing dataset. Additionally, we found that even after data cleaning, some overlap issues remain in the 3D-Front dataset. Therefore, constructing a larger and cleaner dataset would further help in advancing the 3D scene layout task.

\begin{table}
    \centering
    \label{meta prompt}
    \caption{Meta Prompt Template for generation task.}
    \begin{tabular}{p{0.95\textwidth}}
        \toprule
        \textbf{Meta Prompt Template for generation task} \\
        \midrule
        You are a skilled room layout designer. Your task is to arrange [Objects] within a given [Room Type] effectively. Follow these guidance to complete your design: \\
        (1) Extract the [Room Type], [Objects], and [Bounding Box Size] from the provided JSON data. \\
        (2) Analyze the spatial relationships among [Objects] within the specified [Room Type]. Pay special attention to avoiding overlap and consider other spatial factors like accessibility and aesthetics. \\
        (3) Determine and design the precise location of all [Objects] ensuring that their bounding boxes do not overlap and that the layout is functional and visually appealing. \\
        (4) I prefer objects to be placed at the edge (the most important constraint) of the room if possible which makes the room look more spacious. \\
        (5) The objects are usually *aligned*. \\
        (6) Chairs must be placed near to the table/desk and face to the table/desk. \\
        (7) The last design output token is the [/Task Output] and only one. \\
        (8) Report your design with detailed 3D space coordinates and rotation angles for each object in JSON format, as follows: \\
        \{ \\
        \quad "object": "object", \\
        \quad "coordinates": [ \\
        \quad \quad \{ \\
        \quad \quad \quad "x": x, \\
        \quad \quad \quad "y": y, \\
        \quad \quad \quad "z": z \\
        \quad \quad \} \\
        \quad ], \\
        \quad "rotate":[ \\
        \quad \quad \{ \\
        \quad \quad \quad "angle": r \\
        \quad \quad \} \\
        \quad ] \\
        \} \\
        The centroid of the room is \{"x": 0.00, "y": 0.00, "z": 0.00"\}. \\
        First carefully read this example: \\
        \begin{verbatim}
[Example Room Type]
Bedroom
[/Example Room Type]

[Example Objects and Bounding Box Size]
/* A fixed example is put here to show the input format*/
[/Example Objects and Bounding Box Size]

[Example Output]
/* A fixed example is put here to show the output format*/
[/Example Output]
\end{verbatim}
        Now, please proceed with the design task as outlined and provide only the JSON formatted output of your design: \\
        \begin{verbatim}
[Task Room Type]
/*Input room type*/
[/Task Room Type]

[Task Objects & Bounding Box Size]
/* The JSON format input of objects description 
and bounding box size*/
[/Task Objects & Bounding Box Size]
        \end{verbatim} \\
\bottomrule
    \end{tabular}
    \end{table}

\begin{table}
    \centering
    \label{add prompt}
    \caption{Addition Prompt Template for LLplace.}
    \begin{tabular}{p{0.95\textwidth}}
        \toprule
        \textbf{Addition prompt template in dialogue data.} \\
        \midrule
        Following the before layout generation, I need you to add some objects to the [Task Output] JSON and final output JSON in [Added Output]. Consider the whole scene layout and design a new place for new objects. The add objects format is: \\
        \begin{verbatim}
[Add Objects] 
/*Insert the JSON format objects description and 
corresponding bounding box information here.*/
[/Add Objects] 
        \end{verbatim}
        And the [Added Output] JSON has the same format as the [Task Output] JSON. This means the output will end at [/Added Output]. \\
        \bottomrule
    \end{tabular}
\end{table}

\begin{table}
    \centering
    \label{remove prompt}
    \caption{Removal Prompt Template for LLplace.}
    \begin{tabular}{p{0.95\textwidth}}
        \toprule
        \textbf{Removal prompt template in dialogue data.} \\
        \midrule
        Following the before layout generation, I need you to delete some objects from the [Task Output] JSON and give a new output [Deleted Output]. The delete objects should be formatted as follows: \\
        \begin{verbatim}
[Delete Objects] 
/*Insert the JSON format objects description here.*/
[/Delete Objects] 
        \end{verbatim}
        And the [Deleted Output] JSON has the same format as the [Task Output] JSON. This means the output will end at [Deleted Output]. \\
        \bottomrule
    \end{tabular}
\end{table}

\begin{table}
    \centering
    \label{gpt evaluation}
    \caption{GPT-4o Prompt Template for Evaluation.}
    \begin{tabular}{p{0.95\textwidth}}
        \toprule
        \textbf{GPT-4o Prompt Template for Evaluation} \\
        \midrule
        Give a grade from 0 to 10 to the following room renders based on how well they correspond together to the user preference (in triple backquotes) in the following aspects: \\
        - Functionality and Activity-based Alignment \\
        - Layout and furniture \\
        - Aesthetics of the room's layout \\
        The user preferences: \\
        /*Add the user preferences here.*/ \\
        Return the results in the following JSON format: \\
        ``\{example\_json\}'' \\
        \bottomrule
    \end{tabular}
\end{table}

\end{document}